\documentclass[conference]{IEEEtran}
\IEEEoverridecommandlockouts
\usepackage{cite}
\usepackage{amsmath,amssymb,amsfonts}
\usepackage{algorithmic}
\usepackage{graphicx}
\usepackage{textcomp}
\usepackage{xcolor}
\def\BibTeX{{\rm B\kern-.05em{\sc i\kern-.025em b}\kern-.08em
    T\kern-.1667em\lower.7ex\hbox{E}\kern-.125emX}}
\begin{document}

\title{SSGD: A safe and efficient method of gradient
descent\\}

\author{\IEEEauthorblockN{Jinhuan Duan,
Xianxian Li*\thanks{*corresponding author.}, Shiqi Gao,
Jinyan Wang*, Zili Zhong}
\IEEEauthorblockA{\textit{Guangxi Key Lab of Multi-source Information Mining $\&$ Security} \\
\textit{Guangxi Normal University, Guilin, China} \\}
\IEEEauthorblockA{\textit{School of Computer Science and Information Technology} \\
\textit{Guangxi Normal University, Guilin, China} \\
jinhuanduan@gmail.com, lixx@gxnu.edu.cn, shiqigao817@gmail.com, wangjy612@gxnu.edu.cn, zlizhongy@gmail.com}}
\maketitle

\begin{abstract}
    With the vigorous development of artificial intelligence technology, various engineering technology applications have been implemented one after another. The gradient descent method plays an important role in solving various optimization problems, due to its simple structure, good stability and easy implementation. In multi-node machine learning system, the gradients usually need to be shared. Shared gradients are generally unsafe. Attackers can obtain training data simply by knowing the gradient information. In this paper, to prevent gradient leakage while keeping the accuracy of model, we propose the super stochastic gradient descent approach to update parameters by concealing the modulus length of gradient vectors and converting it or them into a unit vector. Furthermore, we analyze the security of super stochastic gradient descent approach. Our algorithm can defend against attacks on the gradient. Experiment results show that our approach is obviously superior to prevalent gradient descent approaches in terms of accuracy, robustness, and adaptability to large-scale batches.
\end{abstract}

\begin{IEEEkeywords}
gradient descent, neural network, data reconstruction attack, differential privacy, distributed machine learning
\end{IEEEkeywords}

\section{Introduction}
Gradient descent (GD) is a technique to minimize an objective function, which is parameterized by the parameters of a model, by updating the parameters with the opposite direction of the gradient of the objective function about the parameters \cite{R16}. It has widely been applied in solving various optimization problems because of its simplicity and impressive generalization ability \cite{YC21}. But it is born with a heart of revealing privacy. Mathematically, the gradient is the parametric derivative of the loss function, which is explicitly calculated from the given training data and its true label. Therefore, an attacker may extract the sensitive information of the original training data from the captured gradients. Recently, researches have shown that an attacker, which captures the gradient of a training sample, can successfully infer its property \cite{MSC+19}, tag \cite{ZMB20}, class representation \cite{HAP17, WSZ+19}, or the data input itself \cite{ZLH19, ZMB20, GBD+20, PZY+20}, with high accuracy. In practical deep learning systems, the gradient of multiple samples is widely used to improve efficiency and performance, which can also be viewed as the per-coordinate average of the single-sample gradients. Is multi-sample gradient safer for the privacy of training data? Unfortunately, Pan et al. \cite{PZY+20} gave the theoretical analysis to indicate that multi-sample gradient still leak samples and labels under certain circumstances.

Since the work of Zhu et al. \cite{ZLH19} was proposed, there is a branch of research \cite{ZLH19, ZMB20, GBD+20, PZY+20} to explore a violent but universal method for successful data reconstruction attacks, and some meaningful empirical results are given on Cifar-10 and ImageNet. These works are based on the same learning-based framework. First, a batch of unknown training samples are used as variables, and then the optimal training samples are searched by minimizing the distance between the ground-truth gradient and the gradient calculated by the variables. The main difference between them is the choice of minimizing distance function. L2 and cosine distances are used in \cite{ZLH19,ZMB20} and \cite{GBD+20}, respectively. Although Zhao et al. \cite{ZMB20} used the properties of neural networks to recover the label of a single sample prior to the learning-based attack, this technique is only suitable to single-point gradient. It is same to \cite{ZLH19} in the multi-sample case. Pan et al. \cite{PZY+20} gave a theoretical explanation of information leakage for single sample in a fully connected neural network with Relu activation function. Furthermore, they showed that there exists leakage of samples and labels for multi-samples under certain circumstances by utilizing the internal information between neurons, and extended the model to ResNet-18 \cite{HZR+16}, VGG-11 \cite{SZ15}, DenseNet-121 \cite{HLM+17}, AlexNet \cite{Kri14}, shufflenet v2-x0-5 \cite{MZZ+18}, InceptionV3 \cite{SVI+16}, GoogLeNet \cite{SLJ+15}, and MobileNet-V2 \cite{SHZ+18}.

To solving the problem of gradient security, Bonawitz et al. \cite{BIK+17} designed a secure aggregation protocol, Phong et al. \cite{PAH+18} encrypted the gradient before sending it, and Abadi et al. \cite{ACG+16} used differential privacy to protect the gradient. However, these three methods have their limitations. Secure aggregation \cite{BIK+17} requires the gradient to be an integer, so it is not compatible with most CNNs. Homomorphic encryption \cite{PAH+18} is only for parameter servers, and differential privacy\cite{ACG+16} protects the gradient while reducing the algorithm's performance. Therefore, this paper proposes a new gradient descent method, super stochastic gradient descent (SSGD) for achieving neuron-level security while maintaining the accuracy of model. Moreover, SSGD has stronger robustness. Phong et al. \cite{PAH+18} analyzed the leakage of the input data in single-layer perceptron with the single-sample and single neuron by using the sigmoid activation function.  Pan et al. \cite{PZY+20} analyzed the leakage of sample data from multi-layer fully connected neural network gradients using the relu activation function, and indicated that multiple samples also reveal privacy. There are two neurons in the last layer which are only activated by the same single sample. Essentially, the leakage is caused by attacking the single-sample gradient. SSGD converts the neuron gradient into a unit vector. This makes that the gradient aggregation of neurons has super randomness. Super randomness may make the performance of the algorithm significantly worse, and it is not easy to converge. We choose to use multiple sample gradient composition updates to increase stability. At the same time, the super randomness also brings strong robustness. the attacker cannot know the true gradient. SSGD invalidates the attack gradient model. Including the attack by searching for the optimal training sample \cite{ZLH19,ZMB20,GBD+20} based on minimizing the distance between the ground-truth gradient and the gradient calculated by the variable, and the attack by solving the equation system \cite{PZY+20} to obtain the training data.

Our contributions are summarized as follows.
\begin{itemize}
    \item  We propose a gradient descent algorithm, called super stochastic gradient descent. The main idea is update the parameters by using the unit gradient vector. In neural networks, neuron parameters are updated by using the unit gradient vector of neurons.
    \item We analyze theoretically that SSGD can realize neuron-level security and defend against attacks on the gradient.
    \item Experimental results show our approach has better accuracy and robustness than prevalent gradient descent approaches.
\end{itemize}

The rest of this paper is organized as follows. In Section 2, we review the basic gradient descent methods and the data leakage by gradients. In Section 3, we describe the super stochastic gradient descent and analyze the safety of our approach. The experimental results are shown in Section 4. Finally, we conclude this paper and give the further work. Validation experiments of data reconstruction attack are shown in the appendix.

\section{Preliminaries}
In this section, we review some basic gradient descent algorithms \cite{R16}, including batch gradient descent (BGD), stochastic gradient descent (SGD) and mini-batch gradient descent (MBGD). The difference among them is that how much data is used to calculate the gradient of the objective function. Then we describe the information leakage caused by gradients \cite{PAH+18}.
\subsection{Basic gradient descent algorithms}
The BGD is a ordinary form of gradient descent, which takes the entire training samples into account to calculate the gradient of the cost function $\ell(\theta)$ about the parameters $\theta$ and then update the parameters by
\begin{equation}
\begin{split}
\theta=\theta-\eta\cdot\nabla_{\theta}\ell(\theta)\label{eq1}
\end{split}
\end{equation}
where $\eta$ is the learning rate. The BGD uses the entire training set in each iteration. Therefore, the update is proceeded in the right direction, and finally BGD is guaranteed to converge to the extreme point.

On the contrary, the SGD considers a training sample $x^{(i)}$ and label $y^{(i)}$ randomly selected from the training set in each iteration to perform the update of parameters by
\begin{equation}
\begin{split}
\theta=\theta-\eta\cdot\nabla_{\theta}\ell(\theta;x^{(i)};y^{(i)})\label{eq2}
\end{split}
\end{equation}

The BGD and SGD are two extremes: one uses all training samples and the other uses one sample for gradient descent. Naturally, their advantages and disadvantages are very prominent. For the training speed, the SGD is very fast, and the BGD can not be satisfactory when the size of training sample set is large. For accuracy, the SGD determines the direction of the gradient with only one sample, resulting in a solution which may not be optimal. For the convergence rate, because the SGD considers one sample in each iteration and the gradient direction changes greatly, it cannot quickly converge to the local optimal solution.

The MBGD is a compromise between BGD and SGD, which performs an update with a randomly sampled mini-batch of $N$ training samples by
\begin{equation}
\begin{split}
\theta=\theta-\eta\cdot\nabla_{\theta}\ell(\theta ;x^{(i:i+N)};y^{(i:i+N)})\label{eq3}
\end{split}
\end{equation}
MBGD decreases the variance of the updates for parameter, so it has more stable convergence. Moreover, the computing of gradient about a mini-batch is very efficient by using highly optimized matrix optimizations existed in advanced deep learning libraries.

\subsection{Analysis of gradient information leakage}
\begin{figure}[h]
\centering
\includegraphics[width=8cm,height=4cm]{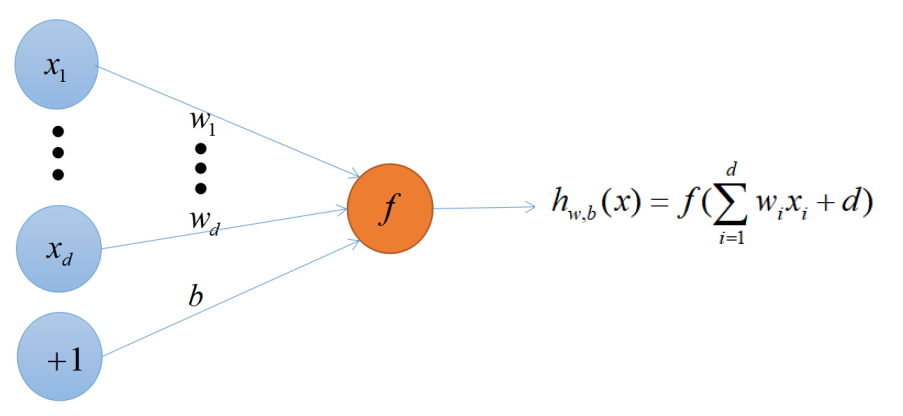}
\caption{Single neuron structure}
\label{fig1}
\end{figure}
 Phong et al. \cite{PAH+18} illustrated that how gradients leak the data information based on a single neuron shown in Fig.~\ref{fig1}. Assume that the $d$-dimensional vector $x\in R^{d}$ represents data input with a label value $y\in R$. $w\in R^{d} $ is the weight parameter and $b\in R$ is the bias, represented uniformly by $\theta={(w,b)}\in R^{d+1}$. $g\in R^{d+1}$ is the gradient vector of the parameter $\theta$. $f$ is an activation function and the loss function $\ell(w, b, x, y)=(h_{w,b}(x)-y)^{2}$, where $h_{w,b}(x)=f(\sum_{i=1}^{d}w_{i}x_{i}+b)$. Let $g=(\sigma_{1},\ldots,\sigma_{k},\ldots,\sigma_{d},\sigma)$, $k\in d$. We have
\begin{equation}
\begin{split}
\sigma_{k}=\frac{\partial\ell(f(x,w,b),y)}{\partial w_{k}}=2(h_{w,b}(x)-y)f'(\sum_{i=1}^{d}w_{i}x_{i}+b)\cdot x_{k}\label{eq4}
\end{split}
\end{equation}
\begin{equation}
\begin{split}
\sigma=\frac{\partial\ell(f(x,w,b),y)}{\partial b}=2(h_{w,b}(x)-y)f'(\sum_{i=1}^{d}w_{i}x_{i}+b)\label{eq5}
\end{split}
\end{equation}
Therefore, we obtain $\sigma_{k}=\sigma\cdot x_{k}$. By solving the system of equations, we can easily get $x$ and $y$. Also, we know that $g$ is determined by $(x,y)$. So $g$ and $(x,y)$ are bijective. In distributed training, usually $w$ and $b$ are parameters that need to be updated and must be known. Then it can infer $(x,y)$ from $g$.
Based on \cite{PZY+20} the single-sample analysis of multi-layer neural networks by using relu activation function, there are also data leakage problem. Although there is no such simple and intuitive leakage of data in a multi-layer neural network, we can still know $x$ and $y$ by analyzing the internal relationship of the neural network and find that $(x, y)$ and $g$ are still bijective.

\section{Super stochastic gradient descent}
In this section, we propose our super stochastic gradient descent approach for preventing gradient leakage while keeping the accuracy, and then analyze in detail the safety of our approach.

\subsection{Our approach}\label{AA}
It was confirmed that the gradient leaks privacy \cite{ZLH19, PAH+18}. For solving the security problem caused by the exchange gradient in stochastic gradient descent or mini-batch gradient descent, we propose the super stochastic gradient descent approach, which can protect the gradient information without losing accuracy by hiding part of the gradient information. The gradient is the first-order partial derivative of the objective function, so it is a vector with both magnitude and direction. We seek the gradient of the objective function to find the fastest descent direction. But it is a little related to the modulus length of the gradient vector. So we hide the modulus length of the gradient vector and convert the gradient vector into a unit vector.

The super-randomness, caused by the aggregation of multiple unit gradient vectors, may lead to poor results. To guarantee that this kind of randomness is friendly, we utilize the following approaches to reduce the uncertainty caused by super-randomness.

For single sample training sample $x^{(i)}$ and label $y^{(i)}$, we use unit gradient vector to update parameter $\theta$
\begin{equation}
\begin{split}
\theta=\theta-\eta\cdot\frac{\nabla_{\theta}\ell(\theta ;x^{(i)};y^{(i)})}{\|\nabla_{\theta}\ell(\theta ;x^{(i)};y^{(i)})\|}\label{eq6}
\end{split}
\end{equation}

For multiple samples, the parameter is updated to
\begin{equation}
\begin{split}
\theta=\theta-\frac{\eta}{m}\cdot\sum_{j=1}^{m}\frac{\nabla_{\theta}\ell(\theta ;x^{(i:i+n)};y^{(i:i+n)})_{j}}{\|\nabla_{\theta}\ell(\theta ;x^{(i:i+n)};y^{(i:i+n)})_{j}\|}\label{eq7}
\end{split}
\end{equation}
where $x^{(i:i+n)}$ represents $n$ samples and $y^{(i:i+n)}$ denotes their labels. The gradient $\nabla_{\theta}\ell(\theta ;x^{(i:i+n)};y^{(i:i+n)})$ of $n$ samples is considered as a basic gradient, and $m$ is the number of basic gradients. Aggregating the unit gradient vectors of $m$ basic gradients on average is to further enhance the stability of the algorithm. The algorithm has higher performance with strong randomness. It is secure to share this unit basic gradient in a distributed environment.

Neuron is the smallest information carrier in the neural network structure. In the neural network, we choose to convert each neuron parameter gradient vector into a unit vector. So the single-layer neural network parameter is updated to
\begin{equation}
\begin{split}
\theta_{k}=\theta_{k}-\frac{\eta}{m}\cdot\sum_{j=1}^{m}\frac{\nabla_{\theta}\ell(\theta ;x^{(i:i+n)};y^{(i:i+n)})_{kj}}{\|\nabla_{\theta}\ell(\theta ;x^{(i:i+n)};y^{(i:i+n)})_{kj}\|}\label{eq8}
\end{split}
\end{equation}
where $\theta_{k}$ represents the $k$-th column or $k$-th row of the parameter matrix in the fully connected layer or convolutional layer (the convolution kernel is regarded as a neuron). In the fully connected layer, $\nabla_{\theta}\ell(\theta ;x^{(i:i+n)};y^{(i:i+n)})_{k}$ is expressed as the $k$-th column of the gradient matrix. And in the convolutional layer, it represents the $k$-th row of the gradient matrix of the convolution kernel. $d$ is the number of rows or columns of the gradient matrix. Therefore, each row or column of the gradient matrix is a unit vector. Then we obtain an average gradient matrix by using $m$ such gradient matrices to update the parameters.

\subsection{The safety of SSGD}
By analyzing the multi-layer neural network with relu activation function on a training sample, the following relationship is obtained in \cite{PZY+20}:
\begin{equation}
\begin{split}
\overline{G}^{i}=\sum_{c}\overline{g_{c}}(D^{i}W^{i-1}\cdots W^{0}X)([W^{H}]_{c}^{T}D^{H}\cdots W^{i+1}D^{i+1})\label{eq9}
\end{split}
\end{equation}
where $X$ is the input data, $\overline{g_{c}}$ represents the $c$-th dimension of the loss vector $\overline{g}$, $D^{i}$ is the activation pattern of the $i$-th neural network, and $\overline{G}^{i}$ and $W^{i}$ denote the gradients and parameters of the $i$-th layer of neural network, respectively.

In fact, the attack gradient models are all solutions to the above equations. The left side of the equation is the $i$-th layer gradient matrix:
\begin{equation}
\begin{split}
\overline{G}^{i}=
\begin{bmatrix}
\sigma_{1,1}^{i} & \sigma_{1,2}^{i} & \ldots & \sigma_{1,n}^{i}\\
\sigma_{2,1}^{i} & \sigma_{2,2}^{i} & \ldots & \sigma_{2,n}^{i}\\
\vdots & \vdots & \ddots & \vdots\\
\sigma_{d,1}^{i} & \sigma_{d,2}^{i} & \ldots & \sigma_{d,n}^{i}\\
\end{bmatrix}_{d\times n}\label{eq10}
\end{split}
\end{equation}
The gradient matrix of our SSGD is :\\
\begin{equation}
\begin{split}
\widehat{G}^{i}=
\begin{bmatrix}
\sigma_{1,1}^{i} & \sigma_{1,2}^{i} & \ldots & \sigma_{1,n}^{i}\\
\sigma_{2,1}^{i} & \sigma_{2,2}^{i} & \ldots & \sigma_{2,n}^{i}\\
\vdots & \vdots & \ddots & \vdots\\
\sigma_{d,1}^{i} & \sigma_{d,2}^{i} & \ldots & \sigma_{d,n}^{i}\\
\end{bmatrix}_{d\times n}
\begin{bmatrix}
\frac{1}{\mu_{1}^{i}} & 0 & \ldots & 0\\
0 & \frac{1}{\mu_{2}^{i}} & \ldots & 0\\
\vdots & \vdots & \ddots & \vdots\\
0 & 0 & \ldots & \frac{1}{\mu_{n}^{i}}\\
\end{bmatrix}_{n\times n}\label{eq11}
\end{split}
\end{equation}
Each column is a unit vector, and $\mu_{k}^{i}$ is the modulus length of the $k$-th column vector of the $i$-th layer gradient matrix, i.e., the modulus length of the $k$-th neuron gradient of the $i$-th layer neural network. Essentially, the parameter matrix of a layer of neural network is multiplied by a diagonal matrix $U^{i}$ on the right, and the value of the diagonal matrix is the reciprocal of the modulus length of the gradient vector of each neuron. By using our SSGD, the \eqref{eq9} is represented as
\begin{equation}
\begin{split}
\widehat{G}^{i}=\sum_{c}\overline{g_{c}}(D^{i}W^{i-1}\cdots W^{0}X)([W^{H}]_{c}^{T}D^{H}\cdots W^{i+1}D^{i+1})U^{i}\label{eq12}
\end{split}
\end{equation}
Where $U^{i}$ is unknown and is not uniquely determined when the loss function are non-convex and non-concave functions. According to \cite{K16}, we know that the loss function of multi-layer neural networks are non-convex and non-concave functions. Due to the dynamicity of $U^{i}$, even if $\overline{g_{c}}$ , $W^{i}$ , and $D^{i}$ are known, $U^{i}$ is unknown. It can not obtain $X$.

Our method hides the correlation between the gradient and the sample, eliminates the information between neurons and achieves neuron-level security. SSGD is a multi-sample training, there is no information leakage problem in \cite{PAH+18}. SSGD can defend against attacks on the gradient.

Since training a model requires rounds of iterations, is it safe to use multiple rounds of iterations? We previously analyzed that the gradient $g$ and the training data $(x,y)$ are bijective in terms of parameter $\theta$, then $g=f(\theta|(x,y))$, where $f$ is a functional relationship. We use $\theta^{i}$ and $\theta^{i+1}$ to denote the training parameters of the $i$-th and $i+1$-th rounds, respectively. Then we have $\theta^{i+1}=\theta^{i}-\eta\cdot g^{i}$. The $i$-th gradient $g^{i}=f(\theta^{i}|(x,y))$ and the $i+1$-th gradient $g^{i+1}=f(\theta^{i+1}|(x,y))$. So we have $\theta^{i+1}=\theta^{i}-\eta\cdot f(\theta^{i}|(x,y))$. Furthermore, we obtian $g^{i+1}=f(\theta^{i}-\eta\cdot f(\theta^{i}|(x,y))|(x,y))$. By compared $g^{i+1}$ with $g^{i}$, we can see that there is not additional information in $g^{i+1}$. The information of the model is only related to the training samples and initial parameters. So the iteration operation does not cause the information leakage.

Can a collusion attack be able to attack super random gradient descent? In an extreme environment, all other parties involved in the training are attackers, and only the local is the attacked. Is it still safe in this case? This problem is equivalent to knowing $\widehat{G}^{i}$ , $U_{n-1}^{i}$, $W$ , $X_{n-1}$ and the model structure, find $x_{n}$. Where $U_{n-1}^{i}$ is a diagonal matrix of $n-1$ dimension.

\begin{equation}
\begin{split}
U^{i}=
\begin{bmatrix}
 U_{n-1}^{i} & 0 \\
  0 &  \frac{1}{\mu_{n}^{i}}\\
\end{bmatrix}_{n\times n}\label{eq13}
\end{split}
\end{equation}
But because of the independence between the samples. The addition of other sample information will not add additional useful information. Substituting the known quantity in \eqref{eq12}, the equation is still $d+1$ unknowns, but only $d$ equations. There is no solution to this equation. Our method can defend against collusion attacks.

\section{Experiments}
Our experiment code has been uploaded to github: https://github.com/duanjinhuan/SSGD. Since Lenet-5 \cite{LBB+98} is a classic convolutional neural network model, we use it to test the performance of our approach by substituting SSGD for basic gradient descents. The Lenet-5 contains two convolutional layers, two pooling layers, and three fully connected layers. The activation function is relu. The input dimensions are 784, and output dimensions are 10. The MINST data set is used for testing the accuracy and robustness of our SSGD approach. It contains 60000 training images and 10000 test images, every image is an 28*28 grayscale image, and each pixel is an octet. The accuracy contains training and test accuracies. We use 60000 training images to train model. The training (test) accuracy is the average value of ten experimental results, and every experiment obtains the average training (test) accuracy of randomly selecting 1000 samples from the training (test) set.

\subsection{Accuracy}
We compare SSGD with SGD, SGDm \cite{Qia99} and Adam \cite{KB15}, which are widely used gradient descent algorithms. The batch size $N$ is set to 16, 32, 64, 128, 256, 512, 1024, 2048, 4096, 8192, and the number of iterations is 10,000. There is not set same learning rate as a good experimental result, because SGD and SGDm have poor adaptability in large batches. The momentum of SGDm is set to 0.999. For the experimental parameters of Adam, we set a good empirical value. The learning rate of Adam is set to 0.005, and the $\beta_{1}$ and $\beta_{2}$ in Adam are set to 0.9 and 0.999, respectively. For SSGD, the learning rate in this experiment is set to 0.1, which is relatively rough. $n$ is set to 1, 4, 8, 16, 32, 64, 128. $m$ is set to 4, 8, 16, 32, 64. When $m$=1, it is the MBGD. The number of iterations is also 10,000.

\begin{table}[htbp]
\small
\caption{The training accuracy of compared algorithms}
\begin{center}
\begin{tabular}{|c|c|c|c|c|}
\hline
$N$  & \textbf{SGD} & \textbf{SGDm} & \textbf{Adam} & \textbf{SSGD}    \\ \hline
  16 & 0.9759$_{(0.0005)}$ & 0.9737$_{(0.0005)}$& 0.9786 &   0.9831   \\ \hline
  32 & 0.9802$_{(0.0005)}$& 0.9776$_{(0.0005)}$&0.9852&  0.9922       \\ \hline
  64 & 0.9932$_{(0.0005)}$& 0.9883$_{(0.0005)}$&0.9913&0.9977       \\ \hline
  128 & 0.981$_{(0.0005)}$&  0.9907$_{(0.0005)}$&0.9949&0.999    \\ \hline
  256 & 0.9959$_{(0.0001)}$& 0.9981$_{(0.0005)}$& 0.9935&0.9992 \\ \hline
  512 & 0.9811$_{(0.0001)}$& 0.9897$_{(0.0005)}$&0.9963&1.0 \\ \hline
  1024 & 0.9831$_{(0.00001)}$& 0.9899$_{(0.0001)}$&0.9994&  1.0   \\ \hline
  2048 & 0.9945$_{(0.00001)}$& 0.9997$_{(0.0001)}$&1.0 & 1.0    \\ \hline
  4096 & 0.9878$_{(0.00001)}$& 0.9907$_{(0.0001)}$& 0.9995&1.0 \\ \hline
  8192 & 0.9961$_{(0.000002)}$& 0.9852$_{(0.0001)}$&0.9999&1.0 \\ \hline
\end{tabular}
\label{tab11}
\end{center}
\end{table}

\begin{table}[htbp]
\small
\caption{The test accuracy of compared algorithms}
\begin{center}
\begin{tabular}{|c|c|c|c|c|}
\hline
$N$  & \textbf{SGD} & \textbf{SGDm} & \textbf{Adam} & \textbf{SSGD}      \\ \hline
16&0.9781 & 0.9778 & 0.9767& $\mathbf{0.9832}$\\ \hline
32& 0.9702 & 0.9768& 0.985&$ \mathbf{0.9876}$ \\ \hline
64& 0.9794& 0.9792& 0.9842& $\mathbf{0.99}$\\ \hline
128& 0.9676& 0.978& 0.9896& $\mathbf{0.9901}$\\ \hline
256& 0.9755& 0.9804& 0.9855& $\mathbf{0.9901}$\\ \hline
512& 0.9665& 0.9789& 0.9814&$\mathbf{0.9888}$ \\ \hline
1024&0.9738& 0.9749& 0.9778& $\mathbf{0.9869}$ \\ \hline
2048& 0.9763& 0.9717& $\mathbf{0.9894}$& 0.9886\\ \hline
4096& 0.9703& 0.9806& 0.9788& $\mathbf{0.9878}$\\ \hline
8192& 0.9785& 0.8994& 0.9753& $\mathbf{0.9855}$\\ \hline

\end{tabular}
\label{tab12}
\end{center}
\end{table}

The comparative experimental results of SGD, SGDm, Adam and SSGD are shown in Table~\ref{tab11} and Table~\ref{tab12}. In Table~\ref{tab11}, the the number in bracket is learning rate. From Table~\ref{tab12}, the performance of our algorithm is better than SGD, SGDm and Adam. And our algorithm has better adaptability to large batches of data. When facing the large batches of data, SGD and SGDm have to reduce the learning rate to adapt to it. And Adam also has obvious overfitting in large batches of data. SSGD has always maintained high precision. On the whole, our algorithm on test accuracy is better and more stable than Adam. Table~\ref{tab13} and Table~\ref{tab14} include the running results our SSGD approach. From Table~\ref{tab13}, we can see that the larger number of training batches ($n\times m$) is, the better the training accuracy is. When the number of training batches is too small, the effect of $n$ on performance is greater than $m$ . For the test accuracy, It is randomly distributed. This may be related to the randomness of the initial parameters and iterations.

\begin{table}[h]
\tiny
\caption{Training accuracy of SSGD}
\begin{center}
\begin{tabular}{|c|c|c|c|c|c|c|c|}
\hline
       & $n=1$ & $n=4$ & $n=8$ &$n=16$ & $n=32$ & $n=64$ & $n=128$  \\ \hline
 $m=4$ & 0.9697& 0.9831& 0.9922& 0.9928& 0.9992& 1.0 & 1.0 \\ \hline
 $m=8$ & 0.9776& 0.9922& 0.9977 & 0.9995 & 1.0 & 1.0 & 1.0 \\ \hline
 $m=16$ & 0.9798& 0.9941& 0.999& 1.0 & 1.0 & 1.0 & 1.0 \\ \hline
 $m=32$ & 0.9867& 0.9963& 0.9992& 1.0 & 1.0 & 1.0  & 1.0 \\ \hline
 $m=64$ & 0.9868& 0.9973&  0.9998& 1.0 & 1.0 & 1.0 & 1.0 \\ \hline

\end{tabular}
\label{tab13}
\end{center}
\end{table}

\begin{table}[h]
\tiny
\caption{Test accuracy of SSGD}
\begin{center}
\begin{tabular}{|c|c|c|c|c|c|c|c|}
\hline
       & $n=1$ & $n=4$ & $n=8$ &$n=16$ & $n=32$ & $n=64$ & $n=128$  \\ \hline
 $m=4$ & 0.9717& 0.9832& 0.9842& 0.9846& 0.9895&0.9886& 0.9888 \\ \hline
 $m=8$ & 0.9815& 0.9876& 0.99 & 0.9884& 0.9855&0.9861& 0.9849 \\ \hline
 $m=16$ & 0.9801& 0.9854& 0.9901& 0.9877& 0.9861&0.9869& 0.9832 \\ \hline
 $m=32$ & 0.9851& 0.9804& 0.9901& 0.9833& 0.9819&0.9867& 0.9828 \\ \hline
 $m=64$ & 0.983&  0.9849& 0.9831& 0.9833& 0.9886&0.9878 & 0.9855 \\ \hline

\end{tabular}
\label{tab14}
\end{center}
\end{table}

\begin{figure}[h]
\centering
\includegraphics[width=7cm,height=5cm]{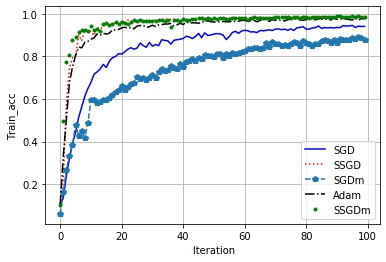}
\caption{The convergence speed of training accuracy}
\label{fig3}
\end{figure}

\begin{figure}[h]
\centering
\includegraphics[width=7cm,height=5cm]{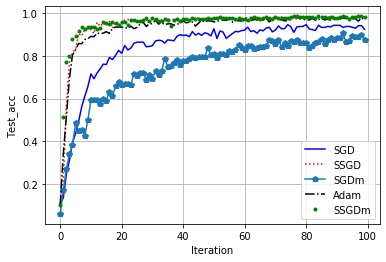}
\caption{The convergence speed of test accuracy}
\label{fig4}
\end{figure}

The convergence speeds about training and test accuracies are shown in Fig.~\ref{fig3} and Fig.~\ref{fig4}, respectively. The value in longitudinal axis is the average accuracy of every 10 iterations. The SSGDm is SSGD with momentum. We choose the intermediate value 256 as the batch number in the convergence experiment, where $n=16$ and $m=16$ for SSGD and SSGDm. The learning rates of SGD and SGDm are 0.0001 and 0.0005, respectively. The momentum of SGDm is set to 0.999. The learning rate of SSGDm is $10/1.0002^t$, where $t$ is the number of iterations, and its momentum is 0.99. The other parameters are consistent with the above experiment. From Fig.~\ref{fig3} and Fig.~\ref{fig4}, we can see that the convergence speed of our algorithm is faster and more stable than SGD , SGDm and Adam.

\subsection{Robustness}
To check the robustness of our algorithm, we add random noise to the gradient. At the same time, we noticed that differential privacy is a way to protect gradient information by adding random noise that meets a certain distribution. To measure the performances of our algorithm and the model with differential privacy to protect gradient. We choose the model in the robustness experiment to add noise that satisfies differential privacy. Compare the performance of the traditional gradient descent algorithm with noise and the performance of ssgd, and the performance of the traditional gradient descent algorithm and ssgd when the same scale of noise is added. In \cite{ACG+16}, the large gradient does not participate in the update, which will seriously affect the gradient descent performance. However, the large gradient participating in the update will cause the noise scale to be too large, making the algorithm effect extremely poor or even unable to converge. Different from cutting gradient value in \cite{ACG+16}, we strictly define sensitivity as the maximum value minus the minimum value in the gradient matrix. We add Laplacian noise of the same scale, and set privacy budget $\epsilon = 4$.

\begin{table}[htbp]
\scriptsize
\caption{Training accuracy of SGDm with different privacy}
\begin{center}
\begin{tabular}{|c|c|c|c|c|c|c|}
\hline
& $n=1$ & $n=4$ & $n=8$ &$n=16$ & $n=32$ & $n=64$       \\ \hline
 $m=4$ & 0.8386& 0.8855& 0.9365& 0.9575& 0.9636&0.9597 \\ \hline
 $m=8$ & 0.8791& 0.9374& 0.9413 & 0.9596 & 0.9533&0.9791  \\ \hline
 $m=16$& 0.8825& 0.9428& 0.9434& 0.9645& 0.9711&0.9785 \\ \hline
 $m=32$& 0.887& 0.927& 0.949& 0.9685& 0.9733&0.9764 \\ \hline
 $m=64$& 0.8775& 0.9248& 0.9523& 0.965& 0.9807&0.9796  \\ \hline

\end{tabular}
\label{tab15}
\end{center}
\end{table}

\begin{table}[htbp]
\scriptsize
\caption{Test accuracy of SGDm with different privacy}
\begin{center}
\begin{tabular}{|c|c|c|c|c|c|c|}
\hline
& $n=1$ & $n=4$ & $n=8$ &$n=16$ & $n=32$ & $n=64$          \\ \hline
$m=4$ & 0.8671& 0.8714& 0.9299& 0.9508&0.9559& 0.9518          \\ \hline
 $m=8$ & 0.8732& 0.9242& 0.9337 & 0.9582& 0.9569&0.9625       \\ \hline
 $m=16$& 0.883& 0.9471& 0.939& 0.9662& 0.9674&0.9699          \\ \hline
 $m=32$& 0.9023& 0.9203& 0.9333& 0.9594& 0.9647&0.9758        \\ \hline
 $m=64$& 0.8765& 0.9163& 0.9514& 0.956& 0.971&0.9678          \\ \hline

\end{tabular}
\label{tab16}
\end{center}
\end{table}
We use SGDm and Adam as the compared algorithms. Also, we have tested SGD algorithm. When $\epsilon = 5$ and the batch number is large, the gradient explosion will occur and the SGD can not be converged. SGDm and Adam algorithms have better robustness. We adjust hyperparameters to get more performance for SGDm and Adam with differential privacy. For SGDm, the learning rate is 0.01, and momentum is 0.99. For Adam, the learning rate is 0.001, $\beta_{1}=0.9$, and $\beta_{2}=0.999$. To make SGDm, Adam and SSGD experiments in the same environment, the batch number is set to $ n\times m$. $n$ is set to 1, 4, 8, 16, 32, 64, and $m$ is set to 4, 8, 16, 32, 64. For each iteration, after the $n$ vectors are added, the Laplace noise of $\epsilon = 4$ that strictly meets the differential privacy is added. The sensitivity is set to the maximum value minus the minimum value of the gradient matrix of the same batch. Then we use SGDm and Adam algorithms to update their parameters, respectively. The number of iterations is 10,000.

\begin{table}[htbp]
\scriptsize
\caption{Training accuracy table of Adam with different privacy}
\begin{center}
\begin{tabular}{|c|c|c|c|c|c|c|}
\hline
 & $n=1$ & $n=4$ & $n=8$ &$n=16$ & $n=32$ & $n=64$        \\ \hline
 $m=4$ & 0.9337& 0.9314& 0.9525& 0.9618& 0.9706&0.9793 \\ \hline
 $m=8$ & 0.9342& 0.9423& 0.9581 & 0.9659 & 0.9768&0.9847  \\ \hline
 $m=16$& 0.9385& 0.9646& 0.9748& 0.9777& 0.9831&0.9936 \\ \hline
 $m=32$& 0.9473& 0.9614& 0.9738& 0.9834& 0.9883&0.9976 \\ \hline
 $m=64$& 0.9543& 0.9627& 0.9824& 0.9897& 0.9968&0.9994  \\ \hline

\end{tabular}
\label{tab17}
\end{center}
\end{table}

\begin{table}[htbp]
\scriptsize
\caption{Test accuracy table of Adam with different privacy}
\begin{center}
\begin{tabular}{|c|c|c|c|c|c|c|}
\hline
 & $n=1$ & $n=4$ & $n=8$ &$n=16$ & $n=32$ & $n=64$               \\ \hline
 $m=4$ & 0.9356& 0.9321& 0.9515& 0.9592& 0.9567 &0.9715         \\ \hline
 $m=8$ & 0.9223& 0.957& 0.9657 & 0.9737& 0.9569&0.9797       \\ \hline
 $m=16$& 0.9481& 0.9607& 0.9606& 0.9723& 0.9726&0.9796          \\ \hline
 $m=32$& 0.9365& 0.958& 0.9693& 0.9658& 0.9763&0.9806        \\ \hline
 $m=64$& 0.9613& 0.9545& 0.9771& 0.9715& 0.9702&0.9833          \\ \hline

\end{tabular}
\label{tab18}
\end{center}
\end{table}
Now, we use SSGDm to verify the robustness of our algorithm. For our algorithm, we use the average of multiple unit gradient vectors to update the gradient. Therefore, the module length of the update gradient vector decreases very slowly, and dynamic learning rates need to be set. The learning rate of SSGDm is set to $10/1.0002^t$, $momentum=0.99$, where $t$ is the number of iterations. We use the same batch settings and the same noise-added methods with $\epsilon = 4$ as SGDm and Adam. Then we use SSGDm to update the parameters. The number of iterations is 10,000.

\begin{table}[htbp]
\scriptsize
\caption{Training accuracy of SSGDm with different privacy}
\begin{center}
\begin{tabular}{|c|c|c|c|c|c|c|}
\hline
 & $n=1$ & $n=4$ & $n=8$ &$n=16$ & $n=32$ & $n=64$         \\ \hline
 $m=4$ & 0.9557& 0.976& 0.9779& 0.9872& 0.9819&0.9904 \\ \hline
 $m=8$ & 0.957& 0.9762& 0.9786 & 0.9858 & 0.9858&0.9936  \\ \hline
 $m=16$& 0.9522& 0.9781& 0.9847& 0.9883& 0.9893&0.9966 \\ \hline
 $m=32$& 0.9559& 0.9807& 0.9846& 0.9911& 0.9927&0.9975 \\ \hline
 $m=64$& 0.9548& 0.9814& 0.9849& 0.9878& 0.9954&0.997  \\ \hline

\end{tabular}
\label{tab19}
\end{center}
\end{table}

\begin{table}[htbp]
\scriptsize
\caption{Test accuracy of SSGDm with different privacy}
\begin{center}
\begin{tabular}{|c|c|c|c|c|c|c|}
\hline
 & $n=1$ & $n=4$ & $n=8$ &$n=16$ & $n=32$ & $n=64$               \\ \hline
 $m=4$  & 0.9552& 0.973& 0.9816& 0.9822& 0.9774&0.9851    \\ \hline
 $m=8$  & 0.9668& 0.9785& 0.9829& 0.979& 0.9832&0.9802   \\ \hline
 $m=16$ & 0.9495& 0.9684& 0.9839& 0.9833& 0.9844&0.986     \\ \hline
 $m=32$ & 0.9594& 0.978& 0.9805& 0.9859& 0.9838&0.9837  \\ \hline
 $m=64$ & 0.9659& 0.9772& 0.9861& 0.9847& 0.9853&0.9885   \\ \hline

\end{tabular}
\label{tab110}
\end{center}
\end{table}
From Table~\ref{tab15} to Table~\ref{tab110}, all three algorithms comply with the law of acquaintance, i.e., the larger the batch size is, the better the accuracy is. We can see that SSGDm is more robust than the Sgdm and Adam algorithms when the same amount of noise is added in training and test accuracy. Compared with SGDm  and Adam, the average test accuracies are increased by $4.12\%$ and $1.60\%$, respectively.

Where is the limit of the robustness of our algorithm? We try to increase the scale of noise and take $\epsilon$ equal to 0.2, 0.5, 1, 2, 4. The experimental environment is the same as the robustness experiment above, and the parameter settings are also the same. The batch number is set to $n=16$ and $m=16$.
\begin{table}[htbp]
\caption{The Training accuracies by varying $\epsilon$}
\begin{center}
\begin{tabular}{|c|c|c|c|c|c|}
\hline
 & $\epsilon=0.2$ & $\epsilon=0.5$ & $\epsilon=1$ &$\epsilon=2$ & $\epsilon=4$      \\ \hline
 SGDm  & & 0.3835& 0.8282& 0.9366& 0.9645    \\ \hline
 Adam  & 0.7937& 0.8755& 0.9131& 0.9499& 0.9777   \\ \hline
 SSGDm & 0.8467& 0.9104& 0.9608& 0.9753& 0.9883    \\ \hline
\end{tabular}
\label{tab111}
\end{center}
\end{table}

\begin{table}[htbp]
\caption{Test accuracies by varying $\epsilon$}
\begin{center}
\begin{tabular}{|c|c|c|c|c|c|}
\hline
 & $ \epsilon=0.2$ & $\epsilon=0.5$ & $\epsilon=1$ &$\epsilon=2$ & $\epsilon=4$     \\ \hline
 SGDm  & &0.3745& 0.8166& 0.9344& 0.9662            \\ \hline
 Adam  & 0.8074& 0.8813& 0.914& 0.9416& 0.9723   \\ \hline
 SSGDm & 0.8481& 0.9395& 0.9671& 0.9719& 0.9833   \\ \hline
\end{tabular}
\label{tab112}
\end{center}
\end{table}

From Table~\ref{tab111} and Table~\ref{tab112}, it is clear that our SSGDm has obvious advantages in robustness. The greater the amount of noise is, the more obvious the robustness is. Compared with Adam and SGDm, the average test accuracies are increased by $3.87\%$ and $19.25\%$, respectively.

\section{Conclusions}
In this paper, we proposes a new gradient descent approach, called super stochastic gradient descent. The SSGD enhances the randomness of gradients to protect against gradient-based attacks. Simultaneously, we use multi-sample aggregation to enhance stability and eliminate the uncertainty brought about by ultra-randomness. Our approach achieves neuron-level security and can defend against attacks on the gradient. Experimental results demonstrate that SSGD has good accuracy and strong robustness because its stability and randomness are enhanced. In the future, we will extend our idea to other machine learning models.
\section*{Acknowledgment}
This paper was supported by the National Natural Science Foundation of China (Nos. 61672176, 61763003 and 61941201), Research Fund of Guangxi Key Lab of Multi-source Information Mining \& Security (No. 19-A-02-01), Innovation Project of Guangxi Graduate Education (No. XYCSZ2020072), Guangxi 1000-Plan of Training Middle-aged/Young Teachers in Higher Education Institutions, Guangxi ``Bagui Scholar'' Teams for Innovation and Research Project, Guangxi Talent Highland Project of Big Data Intelligence and Application, Guangxi Collaborative Innovation Center of Multisource Information Integration and Intelligent Processing.

\bibliographystyle{unsrt}
\bibliography{references}

\section{Appendix}
The experimental code comes from: https://github.com/PatrickZH/Improved-Deep-Leakage-from-Gradients. The following experimental diagrams include the experimental results of DLG\cite{ZLH19} and iDLG\cite{ZMB20} attacking the existing gradient descent method and SSGD algorithm on MINST datasets and CIFAR100 datasets. From the experimental results, we can see that our algorithm can defend against DLG\cite{ZLH19} and iDLG\cite{ZMB20}.
\begin{figure}[htbp]
\centering
\begin{minipage}[t]{0.48\textwidth}
\centering
\includegraphics[width=8cm]{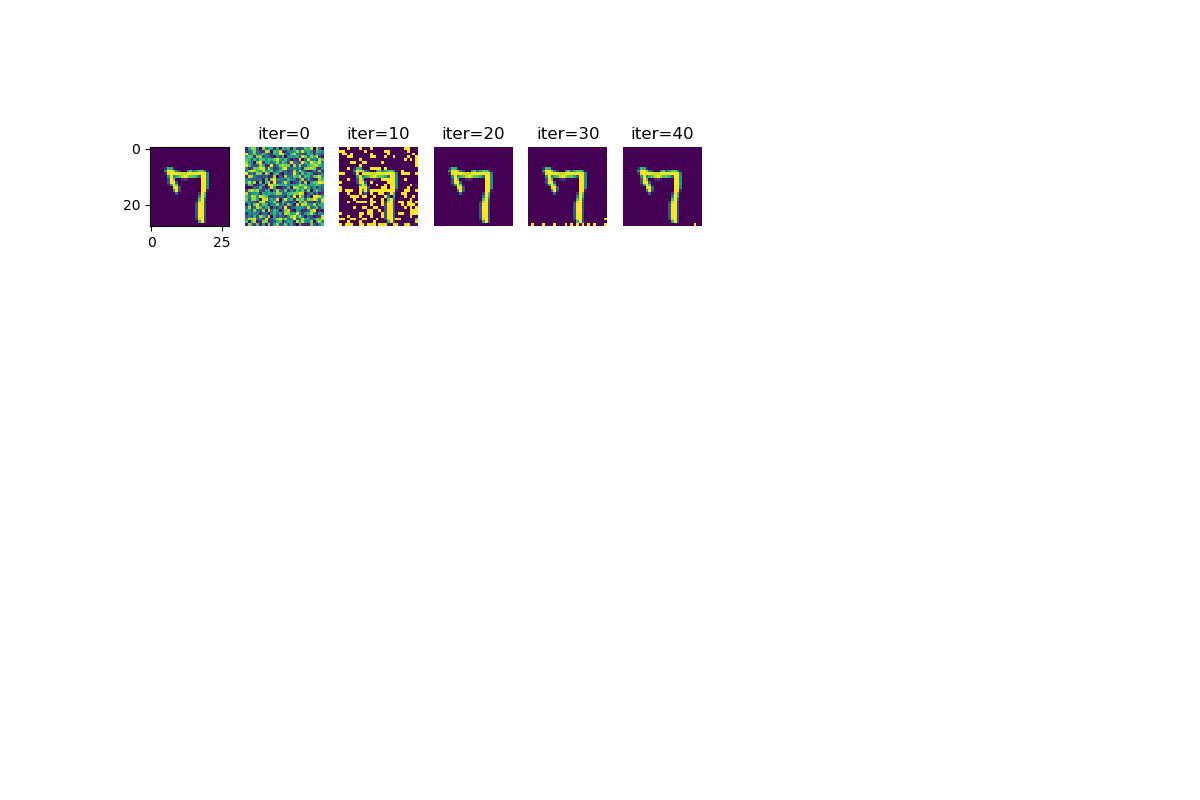}
\caption{DLG on MINST dataset}
\end{minipage}
\begin{minipage}[t]{0.48\textwidth}
\centering
\includegraphics[width=8cm]{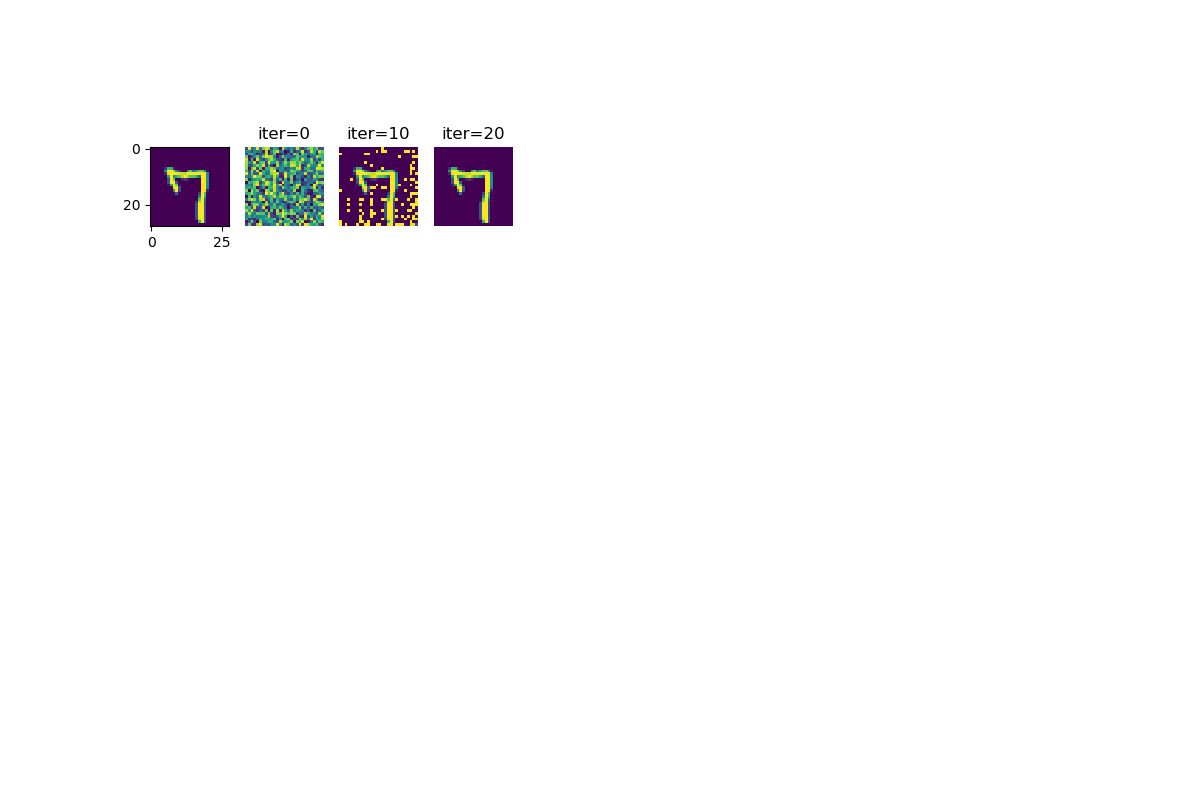}
\caption{iDLG on MINST dataset}
\end{minipage}
\end{figure}

\begin{figure}[htbp]
\centering
\begin{minipage}[t]{0.48\textwidth}
\centering
\includegraphics[width=8cm]{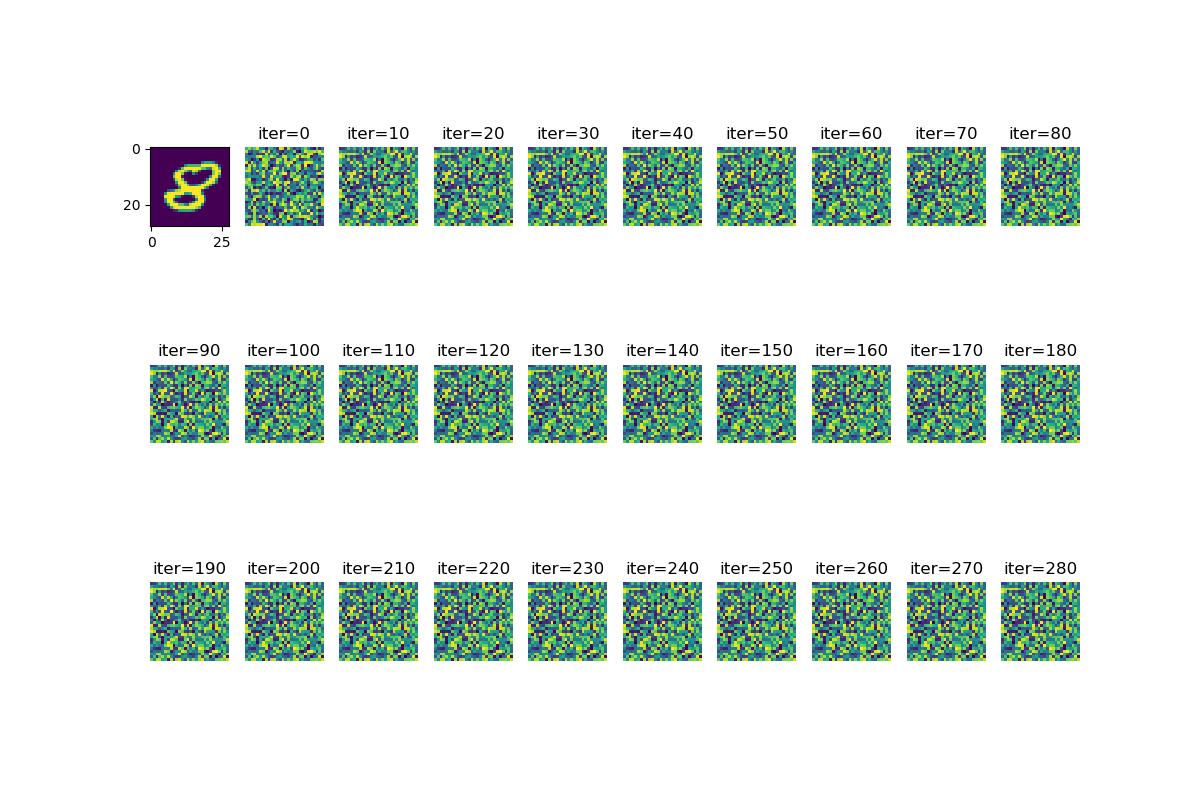}
\caption{DLG with SSGD on MINST dataset}
\end{minipage}
\begin{minipage}[t]{0.48\textwidth}
\centering
\includegraphics[width=8cm]{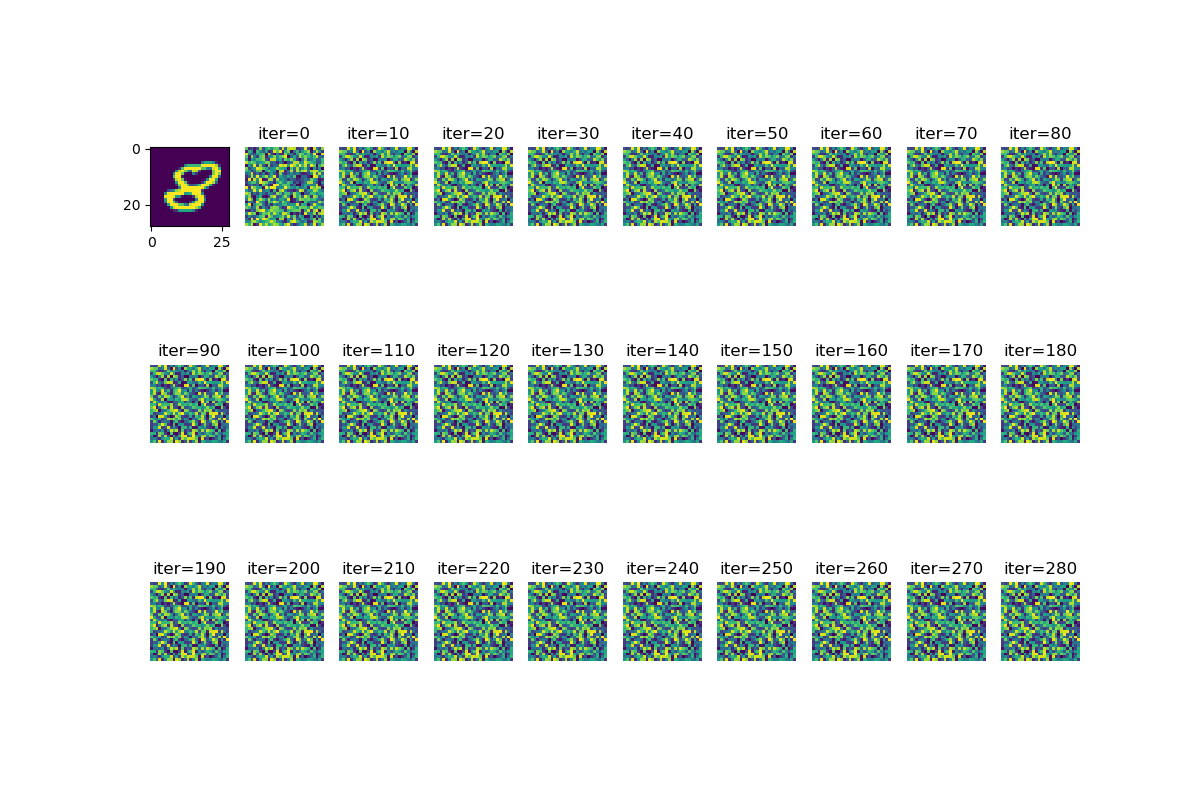}
\caption{iDLG with SSGD on MINST dataset}
\end{minipage}
\end{figure}

\begin{figure}[h]
  \centering
  \includegraphics[width=8cm,height=7cm]{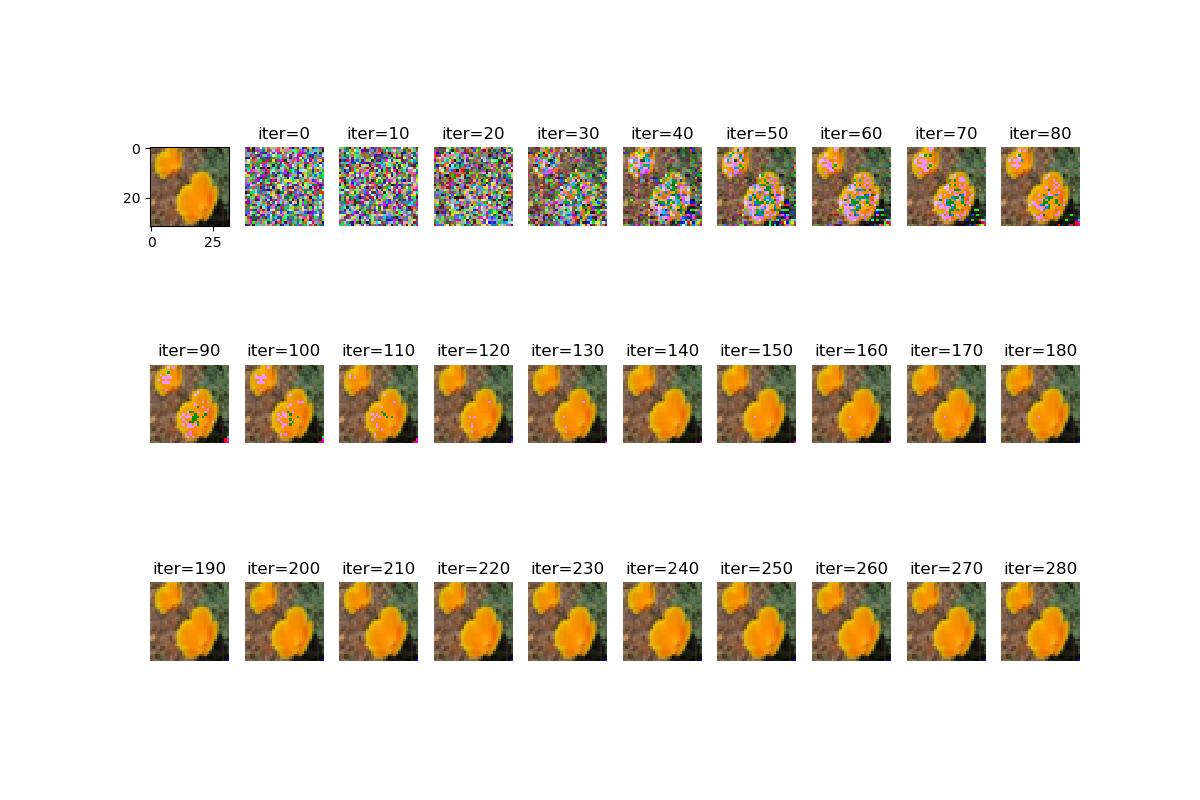}
  \hspace{1in}
  \includegraphics[width=8cm,height=7cm]{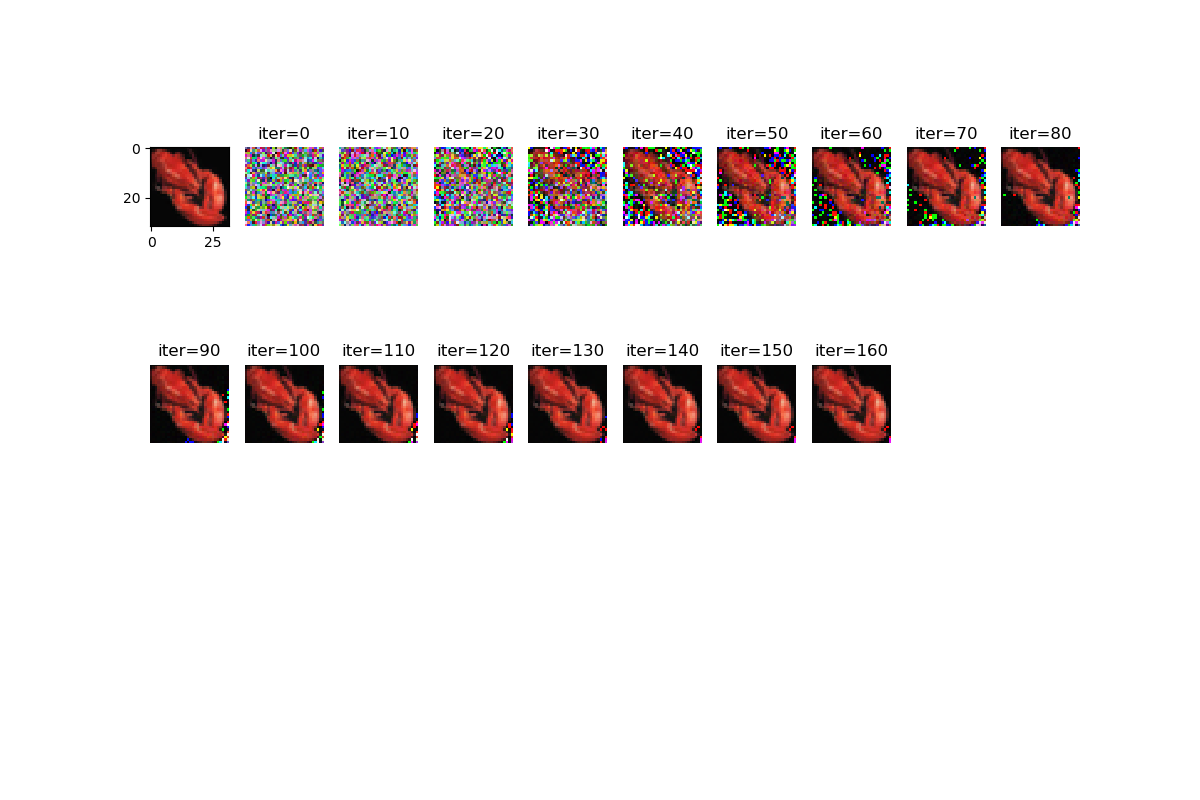}
  \caption{DLG on CIFAR100 dataset}
\end{figure}
\begin{figure}[h]
  \centering
  \includegraphics[width=8cm,height=7cm]{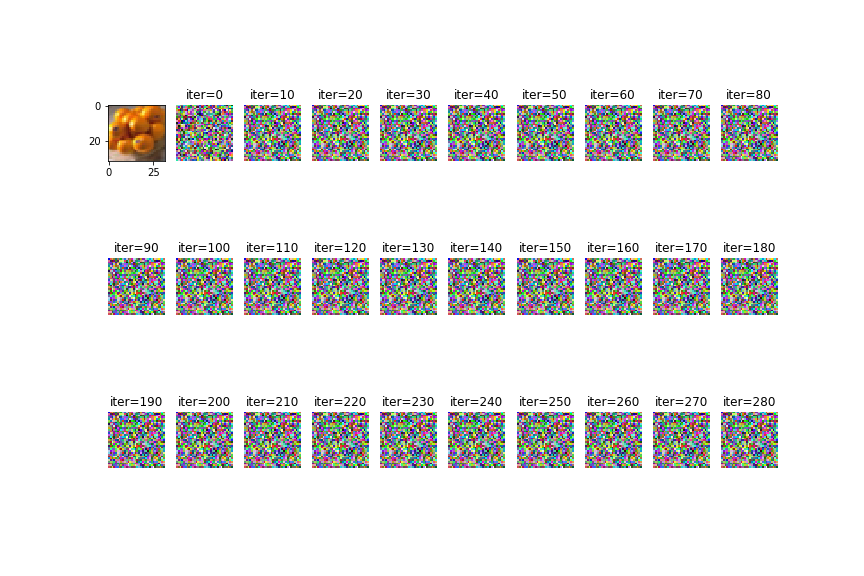}
  \hspace{1in}
  \includegraphics[width=8cm,height=7cm]{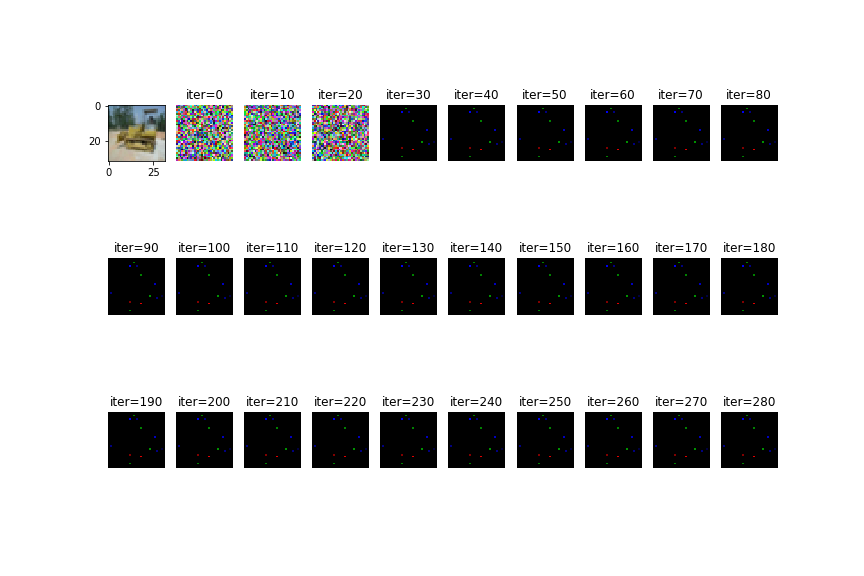}
  \caption{DLG with SSGD on CIFAR100 dataset}
\end{figure}

\begin{figure}[h]
  \centering
  \includegraphics[width=8cm,height=7cm]{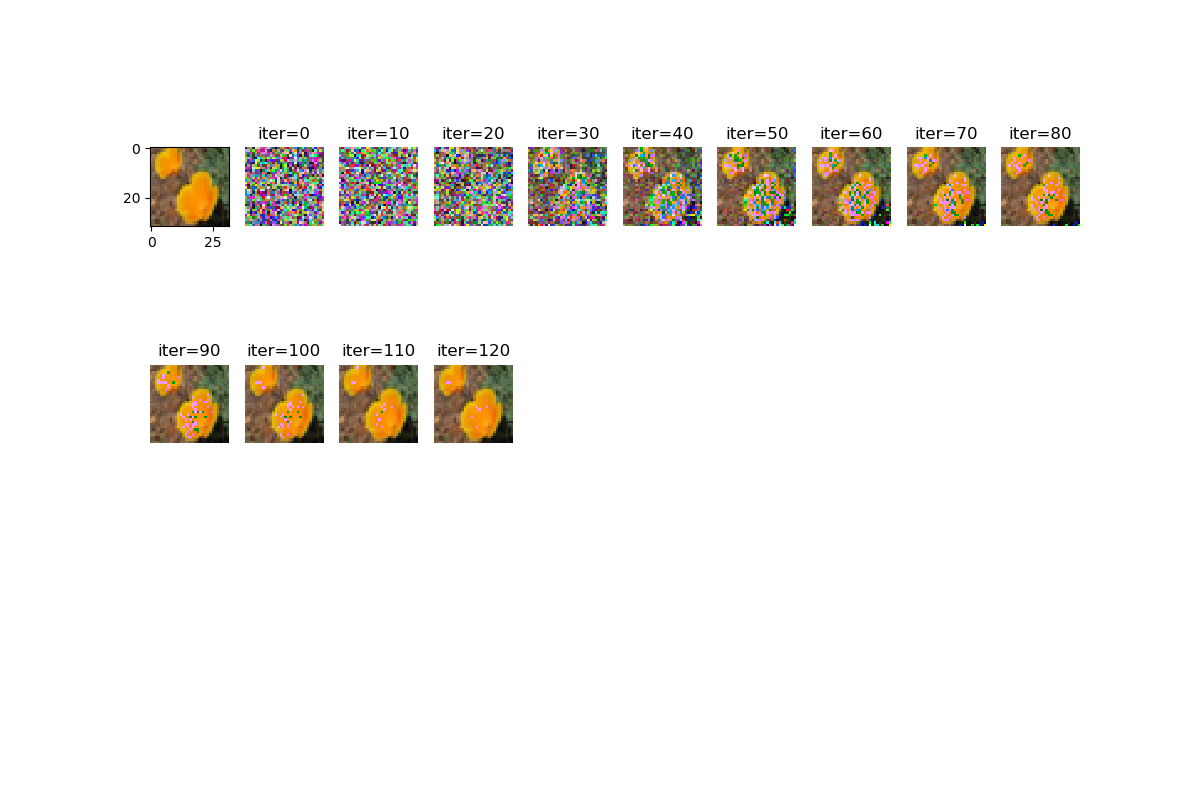}
  \hspace{1in}
  \includegraphics[width=8cm,height=7cm]{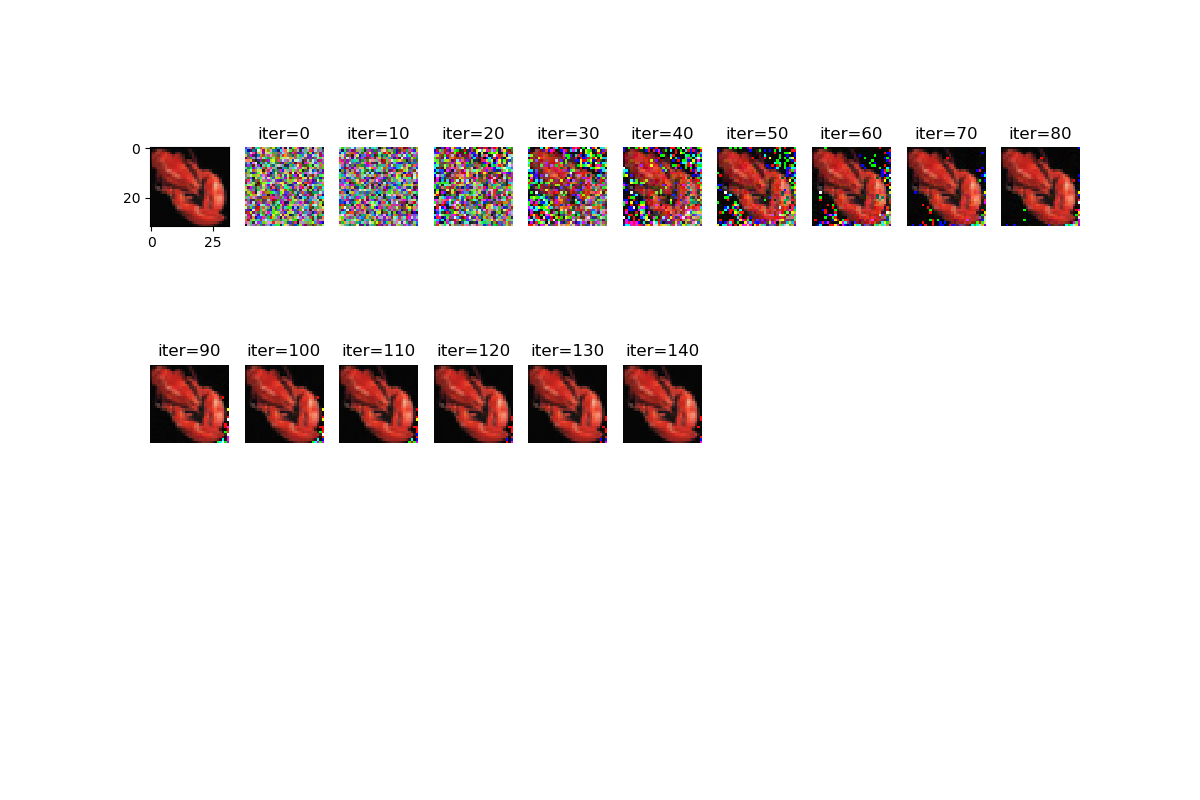}
  \caption{iDLG on CIFAR100 dataset}
\end{figure}
\begin{figure}[h]
  \centering
  \includegraphics[width=8cm,height=7cm]{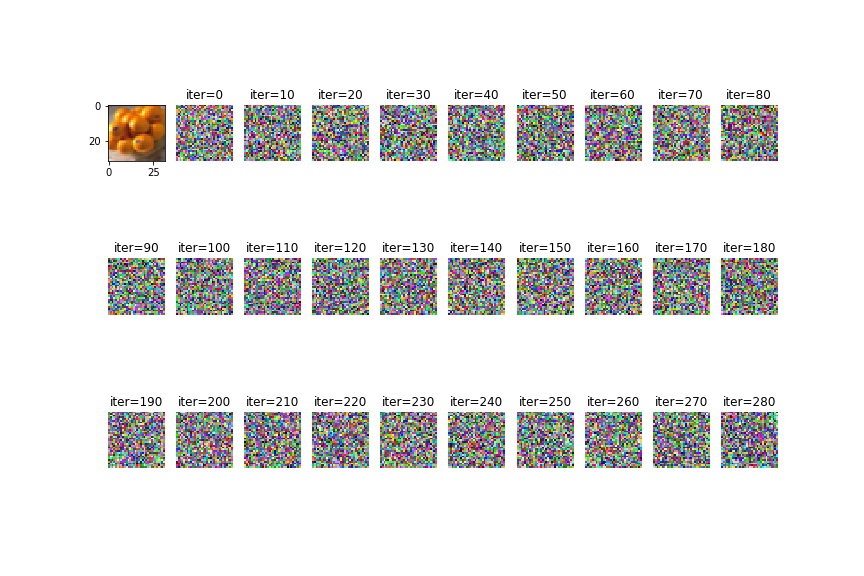}
  \hspace{1in}
  \includegraphics[width=8cm,height=7cm]{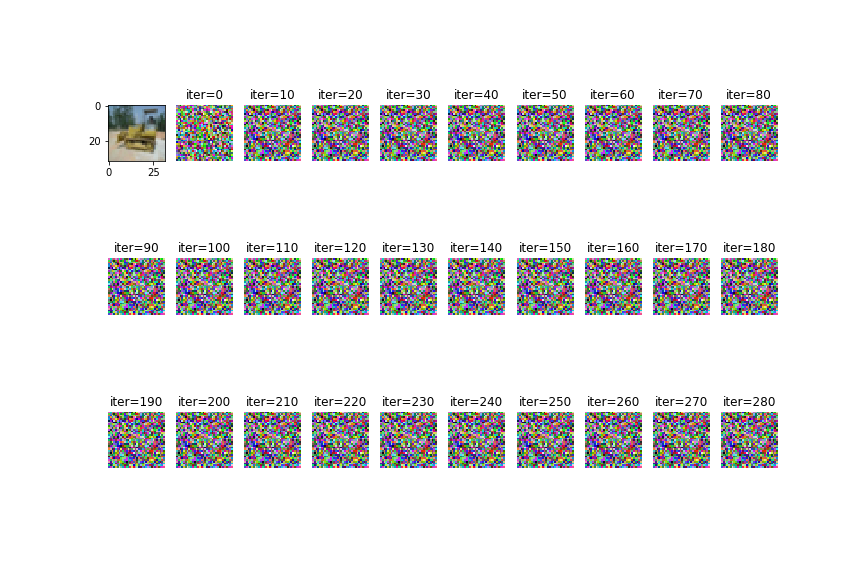}
  \caption{iDLG with SSGD on CIFAR100 dataset}
\end{figure}

\end{document}